\documentclass{llncs}

\usepackage{alltt,url}
\usepackage[english]{babel}

\def\clpfd{CLP($\mathit{FD}$)}

\begin{document}

\mainmatter

\title{Soft Scheduling}

\author{Hana Rudov\'a}

\institute{Faculty of Informatics Masaryk University \\
        Botanick\'a 68a, 602 00 Brno, Czech Republic\\
        e-mail: \email{hanka@fi.muni.cz}} 

\maketitle

\begin{abstract}
Classical notions of disjunctive and cumulative scheduling are studied from
the point of view of soft constraint satisfaction. Soft disjunctive scheduling
is introduced as an~instance of soft CSP and preferences included in this 
problem are applied to generate a~lower bound based on existing discrete
capacity resource. Timetabling problems at Purdue University and Faculty of 
Informatics at Masaryk University considering individual course requirements 
of students demonstrate practical problems which are solved via proposed 
methods. Implementation of general preference constraint solver is discussed 
and first computational results for timetabling problem are presented.
\end{abstract}

\section{Introduction}

Practical solutions of timetabling problems at the Faculty of Informatics and
at Purdue University suggest a~general scheduling problem which may be 
regarded as a~special type of soft constraint satisfaction
problem~\cite{bimonro:semiring97,Ru:thesis01}. Both of the solved
problems are characterized 
by individual requirements of students for a~set of courses they 
would like to attend. This type of problem includes a~large number 
of constraints
due to the
diversity of requirements. Straightforward application of methods from
disjunctive and cumulative scheduling~\cite{baptiste:phd98,Nui:phd94} entails 
an~over-constrained problem. \label{sec-ttp-describe}

Let us consider a~basic timetabling problem which is often solved via the
constraint
programming approach~\cite{Laj:TTPCLP96,GuJuBoi:UniversTTP95,Goltz:UnivTTP98,%
Ru:thesis01}
to demonstrate where possible directions of solution for our problem may lead.
The timetabling problem is represented by given sets of
of course offerings, each consisting of several courses.
Each student enrolls in one or more offerings having some small amount of choice
among elective offerings. Such problem can be
modeled with help of disjunctive scheduling
and global {\tt disjunctive} constraints\footnote{For details about
{\tt disjunctive} (or {\tt serialized}) and {\tt cumulative} constraints see
for example \clpfd\ library of SICStus Prolog~\cite{caotca:clpfdsolver97}.
\label{foo-disj-cumul}}.  One {\tt disjunctive}
constraint expresses the requirement for non-overlapping of courses within one
offering or related offerings
\begin{alltt}
disjunctive([Start1,\ldots,StartN],[Duration1,\ldots,DurationN])
\end{alltt}
where {\tt StartI} and {\tt DurationI} represent starting time and duration of
course {\tt I}. Allocation of courses into the available number of classrooms
is often included in the problem definition via global {\tt cumulative} 
constraint$^{\mathrm{\ref{foo-disj-cumul}}}$
\begin{alltt}
cumulative([Start1,\ldots,StartN],[Duration1,\ldots,DurationN],
           [1,\ldots,1], NbOfRooms)
\end{alltt}
which states that at most {\tt NbOfRooms} courses can be taught at the same
time. Each course needs just one classroom from the available pool of classrooms
which is expressed by the list {\tt [1,\ldots,1]}. 
Let us note that the problem may contain 
several {\tt cumulative} constraints representing classrooms with different 
equipment or size~\cite{Ru:thesis01}.

Representation of the various course separations for each student by the {\tt
disjunctive} constraint would certainly lead into an~over-constrained problem
without any existing solution. Our idea is to propose a~soft version of the {\tt
disjunctive} constraint which would lead to a~minimization of violated student
requirements. Unsatisfied constraints will be handled via preferences
associated with each value in the domain of the
variable. This type of preferences
will be also subsequently used for complementary solving methods in cumulative
scheduling which allow us to reflect the current value of preferences in 
the problem via updating the lower bound. 

This work is based on our earlier research and the
implementation of a~time\-tabling system for Faculty of Informatics described
in~\cite{RuMa:Patat00}. This paper proposed the so called student conflict
minimization problem which is in close relation with the
minimization of unsatisfied constraints considered here. 
Our current intent is to
generalize the earlier proposed approach to be able to extend the problem
solution by adopting other methods for handling preferences. This
extension will be shown in discussion of cumulative scheduling.

\section{Soft Disjunctive Scheduling}

Let us propose a~\emph{soft disjunctive (scheduling) constraint}.
Disjunctive scheduling will be understood in its standard interpretation
as the scheduling of disjunct
activities~\cite{baptiste:phd98,Nui:phd94}. The adjective
soft will express that some of the activities may overlap, as
disjunctive scheduling may result in an~over-constrained problem,
within the context of a~broader problem definition. The degree of
satisfaction of the soft disjunctive constraint
$\mathrm{soft\_disj}(a_1,\ldots a_n)$ wrt.\ 
assignment~$\theta$ of activities $a_1\ldots a_n$ may be expressed by
the number of its pairwise overlapped activities $a_i, a_j$
$$\omega(c\theta) = \sum_{\forall i,j:i < j}  
  \neg \mathrm{disj}(a_i \theta, a_j \theta)$$
where $\mathrm{disj}(a_i \theta, a_j \theta)$ 
evaluates to 1 iff activities $a_i$ and $a_j$ are
disjunctive for assignment $\theta$ of their starting time and duration
variables. Optimal satisfaction of this constraint is then defined as 
the minimum value for $\omega(c\theta)$. In this way, the
$\mathrm{soft\_disj}(a_1,\ldots a_n)$ constraint was transformed
into a~set of soft disjunctive constraints $\mathrm{soft\_disj}(a_i,a_j)$ over
each pair of its activities $a_i,a_j$.
This interpretation also corresponds with MAX-CSP~\cite{FrWa:pcs92}
aimed at satisfying the maximum of constraints\footnote{Our interpretation is
complementary as we need to minimize number of unsatisfied constraints.}.

Let us consider set $C$ of soft disjunctive constraints each defined over some
subset of the set of activities $A$. Taking into account optimal
satisfaction of overall constraint set, we end up with minimization of 
$\sum_{c\in C} \omega(c\theta)$ over instantiations~$\theta$ of activities
in $A$. In the case where particular soft disjunctive constraints share some
activities, each $\mathrm{soft\_disj}(a_i, a_j)$ 
may contribute to the final sum several times. This contribution will
be understood as the weight $w_{ij}$ of a~constraint between activities
$a_i, a_j$. Such
an~interpretation leads to a~weighted CSP~\cite{FrWa:pcs92}, with aim 
to minimize
weighted sum of violated constraints. Satisfaction degree of that problem
corresponds to
\begin{eqnarray}
\min_{\theta} \sum_{\forall a_i,a_j \in A:i < j}
  w_{ij} \times \neg  \mathrm{disj}(a_i \theta, a_j \theta) \enspace .
\label{eqn-weighted}
\end{eqnarray}
where $w_{ij}$ is equal to $0$ if no soft disjunctive constraint between $a_i$
and $a_j$ exists.

For any given assignment $\theta$,
it could be also interesting to consider the number $u(a_i\theta)$ of 
unsatisfied soft disjunctive constraints posted on activity $a_i$
\begin{eqnarray}
u(a_i\theta) = \sum_{\forall a_j \in A: a_j\not=a_i} 
w_{ij} \times \neg  \mathrm{disj}(a_i \theta, a_j \theta) \enspace . 
\label{eqn-weight-one}
\end{eqnarray}
 Such evaluation
would tell us how many overlaps this activity has with other activities.
When activities are relatively equal in importance, we may want to consider
minimization of the worst case unsatisfaction of $u(a_i\theta)$. 
This interpretation will
subsequently lead to a~combination of fuzzy CSP~\cite{DuFarPra:FuzzyCSP94}
and weighted CSP. Let us describe this proposal in the following paragraph.

We will consider $m$ soft disjunctive constraints over set of activities $A$
having cardinality $|A|=n$. Then the maximal value of $u(a\theta)$ corresponds
to $m(n-1)$ because
each activity may occur at most once in each constraint having
maximal number of $(n-1)$ remaining variables as possible candidates for 
overlap.
The set of soft disjunctive constraints included in $u(a_i\theta)$ defines one
fuzzy constraint with a~level of preference corresponding to 
$1-\frac{u(a_i\theta)}{m(n-1)}$. Transformation of $\frac{u(a_i\theta)}{m(n-1)}$
is required to normalize into $\langle 0,1 \rangle$ unit interval.
Computing the complement within unit interval corresponds to
usual interpretation of the level of preference in fuzzy CSP: 
the better the satisfaction of a~fuzzy
constraint is, the closer its value should be to $1$. Finally we may describe 
the best assignments by
\begin{eqnarray}
\max_{\theta} \min_{a_i\in A}  \left[ 1-\frac{u(a_i\theta)}{m(n-1)} \right]
\enspace .
\label{eqn-fuzzy-weighted}
\end{eqnarray}

\subsection{Correspondence with Timetabling Problem}

Let us describe the relationship of evaluation methods proposed 
in Eqns.~\ref{eqn-weighted}
and~\ref{eqn-fuzzy-weighted} with the above presented timetabling problem 
(see Sect.~\ref{sec-ttp-describe}).

First we will consider the weighted CSP interpretation 
from Eqn.~\ref{eqn-weighted}.  The requirement for
non-overlapping of courses for one student
may be taken into account by one soft disjunctive constraint. Summarized
timetabling requests from all students are transformed into a~set of
soft disjunctive constraints over two courses representing activities $a_i$
and $a_j$. The weight $w_{ij}$ of this constraint corresponds to the number of
joint enrollments between courses $a_i$ and $a_j$. Generally the number of
joint enrollments is critical information which is used during
construction of timetables of described shape.

Expression $u(a_i\theta)$ in the second objective 
(Eqn.~\ref{eqn-fuzzy-weighted}) sums the number of students having
requested other course(s) during the expected time of their
enrolled course~$a_i$. This means
that $u(a_i\theta)$ evaluates satisfaction of soft requirements towards the
course $a_i$. Therefore, the overall objective function in
Eqn.~\ref{eqn-fuzzy-weighted} incorporates the
fact that the number of students measured by $u$ should not be ``too bad''
wrt.\ any particular course as it could even force its cancellation.
The disadvantage of this measure is that the proposed expression may take into 
account one student in course~$a_i$ several times due to concurrent scheduling 
of his (her) other courses (at least two other courses).  
However, this inaccuracy of the solution decreases with increasing quality 
of generated timetable and it
could be even abandoned for sufficiently good solutions\footnote{%
Let us expect that at least 90\,\% requirements in course pre-enrollment would
be satisfied. Having average number of courses for each student equal to 10,
it would result in approximately one course per student which he/she is not
able to attend. Taking into account number of time periods within a
week, such average overlapping does not become substantial for
discussed fail contributions.}.

The second objective compares the number of students wrt.\ each course. Such
an~approach could also profit from inclusion of the ratio between violated
requirements per course $u(a_i\theta)$
and the number $s_i$ of enrolled students in course~$a_i$. This type of 
evaluation could be handled in a~similar approach as was proposed
in Eqn.~\ref{eqn-fuzzy-weighted}, i.e.,
by corresponding transformation of the ratio~$\frac{u(a_i\theta)}{s_i}$.

\section{Soft Cumulative Scheduling}
\label{sec-cumul-sched}

The proposed handling of soft disjunctive constraints defines objective
functions with accumulating weights of violated disjunctions over pairs of
activities. This type of preference introduces new information within
the problem which could be also interesting to use in other parts of
problem solution.

Let us consider a~{\tt cumulative} constraint which constrains the
maximal number of
activities to be scheduled at the same time due to discrete capacity
of an~existing resource they all require. We may also need to constrain the
minimal number of activities on that resource (e.g., by set of {\tt atleast}
constraints\footnote{For each {\tt Time}, we can post one 
{\tt atleast(Time, TimeList, Minimum)} constraint expressing that at
least {\tt Minimum} domain variables from {\tt TimeList} has to be equal {\tt
Time}.}). Taking into account only the 
classical hard version of these constraints, 
we are not able to handle available resource space (minimal--maximal usage) 
assigned to the
set of activities by means of preferences included in the problem.
In this section, we would like to propose possible directions for complementary 
solution methods which also reflect
preferences included in problem. This approach will define 
a~\emph{soft cumulative constraint} for one discrete capacity resource.
Let us note that these methods will not change
preferences in the problem as no new violations of soft constraints
are generated. 

First let us define a~model of preferences:~each valid starting 
time of an~activity $a_i$ is associated with a~weight 
expressing how desirable a~given time is\,---\,the smaller the weight is, 
the more desirable the corresponding starting time is.
The objective is to find an~assignment~$\theta$ of activities which 
minimizes the sum
of these weights, i.e., $\min_{\theta}\sum_i w(a_i\theta)$.
Having some set of activities constrained by a~discrete capacity resource, we
can consider weights of its possible candidates to be assigned at each time. 
The basic premise of the
following consideration is that sufficiently ``good'' candidates,
and a~sufficient number of the ``good'' candidates, have to exist 
for this resource at any time. 

Let us expect that we are given some minimal weight $m(a_i)$ for each 
activity~$a_i$. In the simple case, this weight corresponds to $\min_{\theta}
w(a_i\theta)$, but it may be computed (or even approximated) via any other
method. Then the expression $L=\sum_i m(a_i)$ defines the
lower bound of the solution.
Our aim is to compute how any existing discrete capacity resource 
may worsen this lower bound by their accumulated weights.

Let us note that the results of
computation for one discrete resource directly influence value of $m(a_i)$.
These incremented values should be used by all soft cumulative 
constraints sharing corresponding activities.

Our consideration will take into account activities 
requesting unit capacity of a~discrete resource. The requirements for 
a~non-unit capacity could be handled considering the activity number of times
equal to the required resource capacity.

We will consider a~discrete capacity resource defined over a~subset of 
activities
$A'\subseteq A$ during time interval $t \in t_{\min}\ldots t_{\max}$. Its
capacity may vary over time, i.e., we have given
minimal $c^{\min}_t$ and maximal $c^{\max}_t$ capacities of all times $t$.
Let~$\theta_{a_i\leftarrow t}$ 
be an~assignment which assigns starting time $t$ to activity
$a_i\in A'$ with maximal duration $d_i$. Then we will denote 
\begin{tabbing}
\ $d(t,a_i) = \min\{0, \min$\=$\{w(a_i\theta_{a_i\leftarrow s}),$\=
 $[s=(t-d_i-1)\ldots t \,\land\, (t-d_i-1) > t_{\min}] \,\lor$\\ \nopagebreak
\> \> $[s=t_{\min} \ldots t \,\land\, (t-d_i-1) \leq t_{\min}]$\\
\> $\} - m(a_i)\}$
\end{tabbing}
as a~difference between the possible weight contribution of activity $a_i$ at
time $t$ and the minimal expected weight $m(a_i)$ of activity~$a_i$. Weight 
contribution at time $t$
needs to be selected among all weights $w(a_i\theta_{a_i\leftarrow s})$
for such starting times~$s$ of activity
which possibly include processing of $a_i$ at time~$t$.

Now let us order activities $a_i\in A'$ at each time~$t$
such that expression $d(t,a_i)/d_i\leq d(t,a_j)/d_j$ holds
for all activities $a_i,a_j$ having $i<j$. Then value
$L_{\min}=\sum_{t=t_{\min}}^{t_{\max}}\sum_{i=1}^{c^{\min}_t} d(t,a_i)/d_i$
introduces the minimal contribution of
the soft cumulative constraint to the lower bound~$L$. This lower bound increase
includes only the minimal number of activities which should be scheduled 
at given
time~$t$. The importance of such criteria greatly increases when we also have
information about the expected number of activities $c^{\exp}_t$ to be 
scheduled at time~$t$.
This would allow us to consider a~stronger lower bound contribution given by
$L_{\exp}=\sum_{t=t_{\min}}^{t_{\max}}\sum_{i=1}^{c^{\exp}_t} d(t,a_i)/d_i$.

This proposed lower bound does not require constant duration of all
activities as we have considered maximal duration of each activity. However,
the quality of the lower bound decreases if the
duration of activities is over estimated.

\section{Constraint Solver}

Constraint propagation algorithms for the soft scheduling methods discussed 
are implemented as
a~part of the preference constraint solver for timetabling problem at Purdue
University.
This constraint solver is built on top of the \clpfd\ solver of
SICS\-tus Prolog~\cite{caotca:clpfdsolver97} and implemented with help of 
attributed variables. 
An advantage of this implementation consists in possible inclusion of both
hard constraints of \clpfd\ library and soft constraints of a~new preference
constraint solver.

\subsection{Preference Variables}

The constraint solver implemented handles preferences for each value in the
domain of the variable which will be called \emph{preference variable}.
Each preference corresponds to a~natural number with a~value indicating the
degree to which any soft requirement dependent on that value
is violated. Zero preference therefore means complete satisfaction for the 
corresponding value in the domain of the variable. Any higher
value of preference expresses a~degree of violation which would result from
assignment of that value to variable.
Preferences for each value in the domain of the variable may
be initialized by natural number which allows us
to handle initial costs of values in the domain of variables.

\begin{example}
The following predicate creates the preference variable {\tt A}
\begin{verbatim}
pref( A, [7-5, 8-0, 10-0] )
\end{verbatim}
with initial domain containing values {\tt 7}, {\tt 8}, and
{\tt 10} and preferences {\tt 5}, {\tt 0}, and {\tt 0}, resp. It means
that the value {\tt 7} is discouraged wrt.\ other values. 
\end{example}

\subsection{Soft Disjunctive Constraint}

Propagation of preferences for soft disjunctive constraints is ensured via
global constraint
\begin{verbatim}
soft_disjunctive( S_i, D_i, ListS_j, ListD_j, ListW_ij ) ,
\end{verbatim}
where {\tt S\_i} is a~preference variable for starting time of activity $i$, 
{\tt ListS\_j} is a~list of preference
variables for starting time of all activities requesting soft disjunction 
with activity~$i$. {\tt D\_i} and {\tt ListD\_j} represent corresponding
constant\footnote{Possible extension towards durations as domain variables
is discussed at the end of the section.} durations of activities. 
{\tt ListW\_ij} contains the number of soft disjunctive requirements between
activity~$i$ and particular activities from {\tt ListS\_j} ($w_{ij}$ from
Eqn.~\ref{eqn-weighted}). 

Constraint {\tt soft\_disjunctive} is invoked as soon as the preference 
variable~{\tt S\_i} is instantiated. Then we know exactly which values in
the domain of preference variables in {\tt ListS\_j} 
should be discouraged, i.e., the values that overlap with current 
instantiation of {\tt S\_i}.
For these values, preferences are incremented by corresponding weight from
{\tt ListW\_ij}. When some preference variable does not already contain
critical values in its domain no change need to be done. 

Let us expect that $\theta$ assigns value~$v$ to variable for
starting time {\tt S\_i} of activity~$a_i$. During computation,
preference for 
value~$v$ corresponds to part of sum from Eqn.~\ref{eqn-weight-one}
representing contributions of violated soft disjunctive constraints which
already have an~assigned starting time for the second activity. These
contributions are accumulated up to instantiation of starting time~{\tt S\_i}.
The sum of preferences for complete assignment of starting times corresponds to 
the sum
from Eqn.~\ref{eqn-weighted}, representing degree of satisfaction for the
assignment.

Discrimination among multiple activities is handled by inclusion of maximal
allowed value for any $u(a_i\theta)$ (for the definition see 
Eqn.~\ref{eqn-weight-one}). When any partial contribution to
$u(a_i\theta)$ stored in preference variable
exceeds this threshold, corresponding value is filtered from the domain of
variable to disallow such instantiation. In order to search for (sub-optimal)
solution reflecting objective from Eqn.~\ref{eqn-fuzzy-weighted}, maximal 
allowed value could be 
subsequently decreased during search for better solution wrt.\ final 
preferences of last generated solution. The same filtering of values could be
applied in order to exclude a~large ratio of violated student requirements.

Propagation rules for soft disjunctive constraints included only constant
duration activities. Variable duration activities
could be easily incorporated, however. Values of preferences would be 
incremented up to the minimal duration. When the
minimal duration is increased preferences are updated accordingly. This step
would be repeated up to instantiation of domain variable for duration.

\subsection{Soft Cumulative Constraint}

In this section, we would like to discuss basic implementation issues of soft
cumulative constraints. However, their realization is a~major topic 
for future work.

Soft cumulative constraint
\begin{verbatim}
soft_cumulative( ListS_i, ListD_i, ListC_t)
\end{verbatim}
is defined over preference variables for starting times {\tt ListS\_i},
durations {\tt ListD\_i} of activities, and constant capacities 
{\tt ListC\_t}
representing expected use of discrete capacity resource at each time.

Section~\ref{sec-cumul-sched} proposed a~method for computing lower bound of
solution based on preferences associated with each value in the domain of
variable. This lower bound can be updated during the search for a~solution of
the problem wrt.\ removed values from the domain of variables and increased
values of preferences by soft disjunctive constraints. Finally this
lower bound may be
used to prune the search space when current partial assignment exceeds it.

During computation preference variables from {\tt ListS\_i} are subsequently
instantiated. This fact needs to be reflected by removing newly instantiated
preference variable from lists {\tt ListS\_i} and {\tt ListD\_i} and by
decreasing capacity in {\tt ListC\_t} for all times when activity is processed.

Because each soft cumulative constraint may increase the
minimal weight $m(a_i)$ of
activity $i$ from {\tt ListS\_i} (see Sect.~\ref{sec-cumul-sched}), 
we need to share minimal weights for all
activities among all existing soft cumulative constraints.

\subsection{Search}
\label{sec-search}

As a~part of constraint solver,
we have implemented anytime branch and bound algorithm\,---\,user may specify
time limit of computation or interrupt the optimization and request the
currently best solution. An~additional
value ordering applies preferences via preferred-first strategy, i.e., values
with the best preferences are selected first.
From an~optimistic point,
this could be a~value violating the smallest number of constraints. 
A~new variable ordering heuristic based on preferences
selects the most constrained variable first. However,
the measure is based on the number of soft constraints suspended on the
variable. Ties are broken by selecting variable having the best
preference. 

Let us note that the most constrained heuristics currently corresponds with
selection of a~course having the largest number of joint enrollments with other
courses.

\section{First Computational Results}

Let us present our first computational results 
based on real data from large lecture timetabling problem at Purdue University.
Data set includes 258~courses and 35~classrooms with average classroom
occupation about 74\,\%. Problem definition contains data for joint 
enrollments between courses and basic faculty preferences for
starting time of courses. In the future, problem definition should be extended
by constraints for multi-hour courses,  more
detailed capacity constraints, and additional faculty preferences.

Problem solution includes soft disjunctive constraints and initial costs for
values in domains of
preference variables. Cumulative scheduling is included via hard
constraints only. A~solution was computed by search method described in
Sect.~\ref{sec-search}. First solution was found during 15 seconds on
a~Pentium III/933\,MHz PC violating less than 1\,\% of requirements from joint
enrollment matrix and 1\,\% of requirements given by 
initial preferences. Quality of generated
solution was slightly improved during the following runs of branch and bound
search. However, this increase was not significant. This is probably a
consequence of the high satisfaction degree of the first generated solution.

\section{Conclusions and Future Work}
\label{sec-future}

We have proposed general instances of soft disjunctive and soft cumulative
scheduling constraints based on real timetabling problems at Purdue 
University and at the Faculty of Informatics. Soft disjunctive constraint was
decomposed into a~set of soft disjunctive constraints over pairs of activities.
This decomposition allows us to handle problem via weighted CSP
and combined fuzzy and weighted CSPs. Preferences assigned to each
value of a~variable were considered to handle a~discrete capacity resource.
It was shown how each discrete capacity resource contributes to a~lower bound 
on the solution by preferences place on its activities. 

A~soft constraint solver for handling preferences of each value in the domain
variable was implemented in SICStus Prolog. Current implementation includes
propagation rules for soft disjunctive constraints and a~basic search method
applying preferences. Solver  extension by soft cumulative constraint was
discussed.

First computational results were presented for timetabling problem with
requirements of individual students for a~set of courses. Because the quality 
of the first generated solution was very high additional search of solution space
does not improve solution significantly. We expect that this situation 
will be changed for extended problem definition.
Initial preferences will 
increase and the set of hard constraints will be enlarged.
 We also intend to
handle problems containing up to 800 courses.

Our future work will include further improvement of propagation rules for soft
disjunctive scheduling and implementation of proposed soft cumulative
constraint. Computing the lower bound by soft cumulative constraint does
not consider any relation between activities scheduled at different times.
This could introduce possible directions for further improvements of computed
lower bound. 

We would also like to apply a~preference solver to compute a~new solution based
on existing solution such that
their distance is minimized. Aim of this methods
will be incremental change of generated solution. This would allow us to 
reflect changes in the problem definition without critical changes of
solution.

\section*{Acknowledgements}

This work is supported by the Grant Agency of Czech
Republic under the contract~201/01/0942. 
I would like to thank Keith Murray and Carol Horan for discussion about
timetabling problem at Purdue University.


\begin{thebibliography}{10}

\bibitem{baptiste:phd98}
Philippe Baptiste.
\newblock {\em A~Theoretical and Experimental Study of Resource Constraint
  Propagation}.
\newblock PhD thesis, University of Compi\`{e}gne, 1998.

\bibitem{bimonro:semiring97}
Stefano Bistarelli, Ugo Montanari, and Francesca Rossi.
\newblock Semiring-based constraint solving and optimization.
\newblock {\em Journal of {ACM}}, 44(2):201--236, March 1997.

\bibitem{caotca:clpfdsolver97}
Mats Carlsson, Greger Ottosson, and Bj\"{o}rn Carlson.
\newblock An~open-ended finite domain constraint solver.
\newblock In {\em Programming Languages: Implementations, Logics, and
  Programming}. Springer-Verlag LNCS 1292, 1997.

\bibitem{DuFarPra:FuzzyCSP94}
Didier Dubois, H\'el\`ene Fargier, and Henri Prade.
\newblock Propagation and satisfaction of flexible constraints.
\newblock In {\em Fuzzy Sets, Neural Networks and Soft Computing}, pages
  166--187. Van Nostrand Reinhold, New York, 1994.

\bibitem{FrWa:pcs92}
Eugene~C. Freuder and Richard~J. Wallace.
\newblock Partial constraint satisfaction.
\newblock {\em Artificial Intelligence}, 58:21--70, 1992.

\bibitem{Goltz:UnivTTP98}
Hans-Joachim Goltz, Georg K{\"{u}}chler, and Dirk Matzke.
\newblock Constraint-based timetabling for universities.
\newblock In {\em Proceedings {INAP}'98, 11th International Conference on
  Applications of Prolog}, pages 75--80, 1998.

\bibitem{GuJuBoi:UniversTTP95}
Christelle Gu\'eret, Narendra Jussien, Patrice Boizumault, and Christian Prins.
\newblock Building university timetables using constraint logic programming.
\newblock In Edmund Burke and Peter Ross, editors, {\em Practice and Theory of
  Automated Timetabling}, pages 130--145. Springer-Verlag LNCS 1153, 1996.

\bibitem{Laj:TTPCLP96}
Gyuri Lajos.
\newblock Complete university modular timetabling using constraint logic
  programming.
\newblock In Edmund Burke and Peter Ross, editors, {\em Practice and Theory of
  Automated Timetabling}, pages 146--161. Springer-Verlag LNCS 1153, 1996.

\bibitem{Nui:phd94}
Wilhelmus P.~M. Nuijten.
\newblock {\em Time and resource constrained scheduling: a~constraint
  satisfaction approach}.
\newblock PhD thesis, Eindhoven University of Technology, 1994.

\bibitem{Ru:thesis01}
Hana Rudov\'a.
\newblock {\em Constraint Satisfaction with Preferences}.
\newblock PhD thesis, Faculty of Informatics Masaryk University, 2001.
\newblock See \url{http://www.fi.muni.cz/~hanka/phd.html}.

\bibitem{RuMa:Patat00}
Hana Rudov\'{a} and Lud\v{e}k Matyska.
\newblock Constraint-based timetabling with student schedules.
\newblock In Edmund Burke and Wilhelm Erben, editors, {\em PATAT
  2000\,---\,Proceedings of the 3rd international conference on the {P}ractice
  {A}nd {T}heory of {A}utomated {T}imetabling}, pages 109--123, 2000.

\end{thebibliography}
\end{document}